\documentclass[runningheads]{llncs}
\usepackage{graphicx}
\usepackage{subfig}
\usepackage[utf8]{inputenc}
\usepackage{algorithm}
\usepackage{algpseudocode}
\usepackage{adjustbox}    
\begin{document}
\title{Multimodal 3D Object Detection from Simulated Pretraining}
%
%
\author{Åsmund Brekke\orcidID{0000-0003-1114-5320} \and
Fredrik Vatsendvik\orcidID{0000-0002-5300-0202} \and Frank Lindseth \orcidID{0000-0002-4979-9218}}
\authorrunning{Brekke et.al}
%
\institute{Norwegian University of Science and Technology, Trondheim, Norway
\email{\{aasmunhb,fredrva\}@stud.ntnu.no, frankl@ntnu.no}}
\maketitle              
%
\begin{abstract} 
The need for simulated data in autonomous driving applications has become increasingly important, both for validation of pretrained models and for training new models. In order for these models to generalize to real-world applications, it is critical that the underlying dataset contains a variety of driving scenarios and that simulated sensor readings closely mimics real-world sensors.  
We present the Carla Automated Dataset Extraction Tool (CADET), a novel tool for generating training data from the CARLA simulator to be used in autonomous driving research. The tool is able to export high-quality, synchronized LIDAR and camera data with object annotations, and offers configuration to accurately reflect a real-life sensor array. Furthermore, we use this tool to generate a dataset consisting of 10 000 samples and use this dataset in order to train the 3D object detection network AVOD-FPN, with finetuning on the KITTI dataset in order to evaluate the potential for effective pretraining. 
We also present two novel LIDAR feature map configurations in Bird's Eye View for use with AVOD-FPN that can be easily modified. These configurations are tested on the KITTI and CADET datasets in order to evaluate their performance as well as the usability of the simulated dataset for pretraining. Although insufficient to fully replace the use of real world data, and generally not able to exceed the performance of systems fully trained on real data, our results indicate that simulated data can considerably reduce the amount of training on real data required to achieve satisfactory levels of accuracy.


\keywords{Autonomous driving \and Simulated data \and 3D object detection \and CARLA \and KITTI \and AVOD-FPN \and LIDAR \and Sensor fusion}
\end{abstract}
%
%
%
\section{Introduction}
Machine learning models are becoming increasingly complex, with deeper architectures and a rapid increase in the number of parameters. The expressive power of such models allow for more possibilities than ever before, but require large amounts of labeled data to properly train. 
Labeling of data in the autonomous driving domain requires extensive amounts of manual labour, either in the form of actively producing annotations such as class labels, bounding boxes and semantic segmentation by hand, or by supervising and adjusting automated generation of these using a pretrained ensemble of models from previously labeled data.
For the use of modern sensors such as LIDAR, not many sizable labeled datasets exists, and those that do generally offer little variation in terms of environments or weather conditions to properly allow for generalization to real world conditions. Popular datasets such as KITTI~\cite{kitti_dataset} offers a large array of sensors, but with largely unchanging weather conditions and lighting, while the larger and more diverse BDD100K~\cite{bdd100k} dataset does not include multimodal sensor data, only offering camera and GPS/IMU. The possibility of new sensors being introduced that greatly impact autonomous driving also carry the risk of invalidating the use of existing datasets for training state-of-the-art solutions. 



\subsection{Simulated Data for Autonomous Driving}
With the advances in recent years in the field of computer graphics, both in terms of photorealism and accelerated computation, simulation has been a vital method of validating autonomous models in unseen environments due to the efficiency of generating different scenarios \cite{simulation_testing}. More recently there has been added interest in the use of modern simulators for also generating the data used to train models for autonomous vehicles, both for perception and end-to-end reinforcement learning \cite{prior_3d_maps,carla_end_to_end_imitation,effective_synthetic_data}. There are several advantages in generating training data through simulation.
Large datasets, with diverse conditions can be quickly generated provided enough computational resources, while labeling can be fully automated with little need for supervision. Specific, difficult scenarios can be more easily constructed and advanced sensors can be added provided they have been accurately modeled. Systems such as the NVIDIA Drive Constellation~\cite{nvidiadrive_constellation} are pushing the boundaries for photorealistic simulation for autonomous driving using clusters of powerful NVIDIA GPUs, but is currently only available to automakers, startups and selected research institutions using NVIDIAs Drive Pegasus AI car computer, and only offers validation of models, not data generation for training. However, open source solutions based on state-of-the-art game engines such as Unreal Engine 4 and Unity, are currently in active development and offer a range of features enabling anyone to generate high quality simulations for autonomous driving. Notable examples include CARLA~\cite{carla_sim} and AirSim~\cite{airsim}, the former of which was used for this research.


\section{Simulation Toolkit}
\begin{figure}
    \centering
    \includegraphics[width=\textwidth]{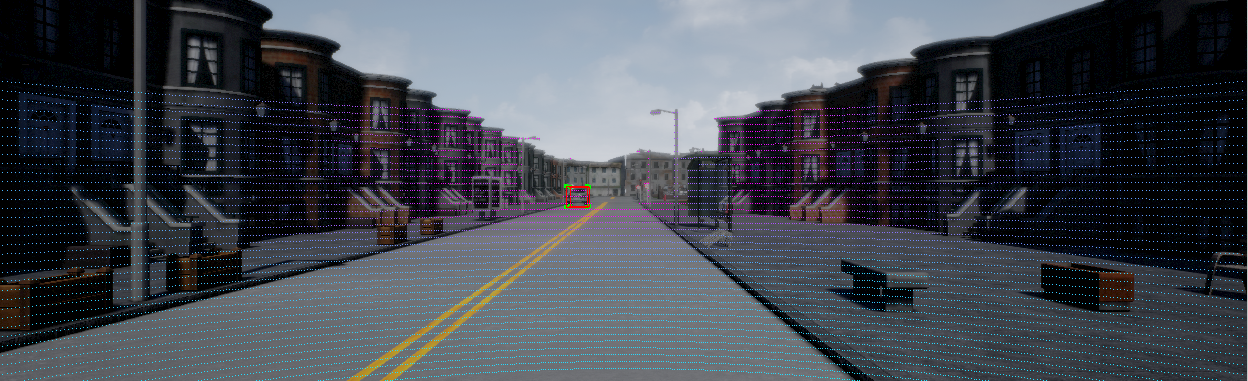}
    \caption{LIDAR point cloud projected to image space. The color of each LIDAR point is determined by its depth value.}
    \label{fig:lidar_projection}
\end{figure}

\begin{figure}%
    \centering
    \subfloat{{\includegraphics[width=\textwidth]{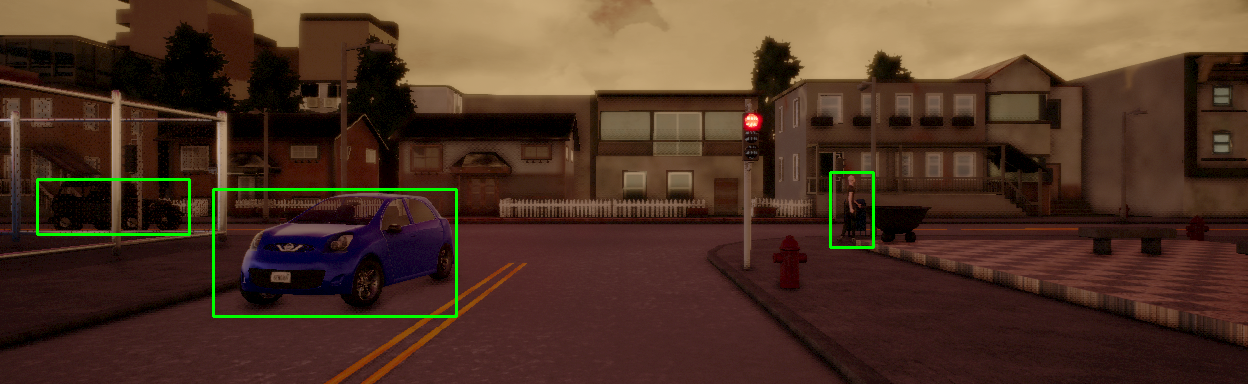}}}%
    \qquad
    \subfloat{{\includegraphics[width=\textwidth]{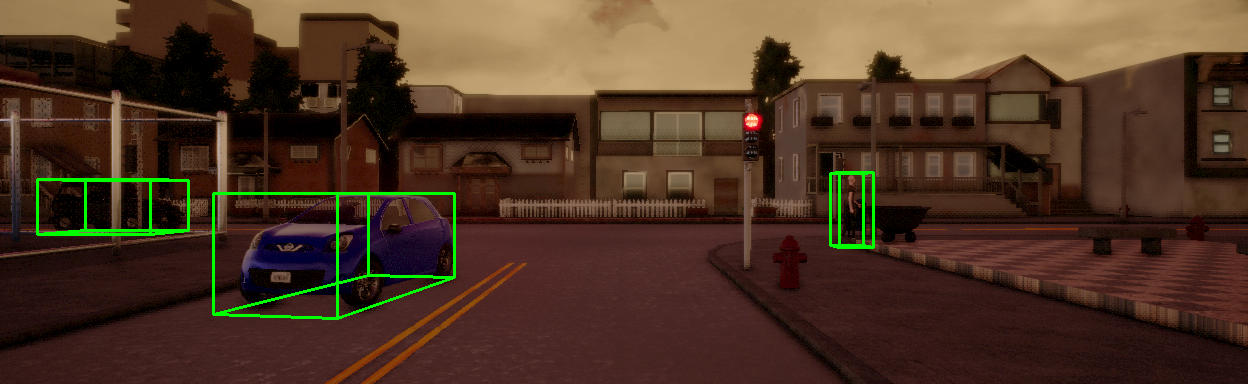}}}%
    \caption{2D (top) and 3D (bottom) bounding boxes as generated by CADET. Class labels are omitted.}%
    \label{fig:bboxes}%
\end{figure}

\begin{figure}
    \centering
    \includegraphics[width=\textwidth]{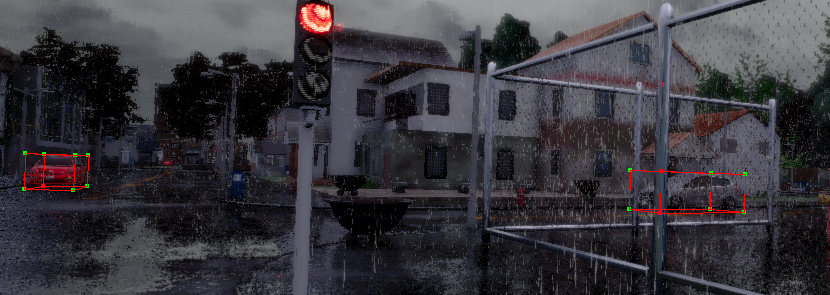}
    \caption{Occluded vertices behind a chain fence. Note that both cars are visible, and thus have a bounding box drawn around them. Occluded and visible vertices are drawn with red and green, respectively.}
    \label{fig:occlusion_carla}
\end{figure}

In order to facilitate the training and validation of machine learning models for autonomous driving using simulated data, the authors introduce the Carla Automated Dataset Extraction Tool (CADET), an open-source tool for generating labeled data for autonomous driving models, compatible with Carla 0.8. The tool supports various functionality including LIDAR to camera projection (Figure \ref{fig:lidar_projection}), generation of 2D and 3D bounding box labels for cars and pedestrians (Figure \ref{fig:bboxes}), detection of partially occluded objects (Figure \ref{fig:occlusion_carla}), and generation of sensor data including LIDAR, camera and groundplane estimation, as well as sensor calibration matrices. All labels and calibration matrices are stored in the data format defined by Geiger et al.~\cite{kitti_dataset}, which makes it compatible with a number of existing models for object detection and segmentation.
As a varied dataset is crucial for a machine learning model to generalize from a simulated environment to real-life scenarios, the data generation tool includes a number of measures to ensure variety. Most importantly, the tool resets the environment after a fixed number of samples generated. Here, a sample is defined as the tuple containing a reading from each sensor, corresponding ground truth labels and calibration data. Resetting the environment entails randomization of vehicle models, spawn positions, weather conditions and maps, and ensures a uniform distribution of weather types, agent models for both pedestrians and cars and starting positions of all vehicles.
The LIDAR and camera sensors are positioned identically and synchronized such that a full LIDAR rotation exists for each image\footnote{Note that this only implies an approximate correspondance between points from the camera and LIDAR sensors.}. The raw sensor data is projected to a unified coordinate system used in Unreal Engine 4 before determining visible objects in the scene, and projecting to the relative coordinate spaces used in KITTI. As the initial LIDAR configuration in CARLA ignores the pitch and roll of the vehicle it is attached to, additional transformations are applied after projection such that the sensor data is properly aligned.
One challenge when generating object labels is determining the visible objects in the current scene.
In order to detect occluded objects, the CARLA \textit{depth map} is utilized. A vertex is defined as occluded if the value of one of its neighbouring pixels in the depth map is closer than the vertex distance to the camera. An object is defined as occluded if at least four out of its eight bounding box vertices are occluded. This occlusion detection performs satisfactory and much faster than tracing the whole object, even when objects are localized behind see-through objects such as chain-link fences, shown in Figure \ref{fig:occlusion_carla}. A more robust occlusion detection can be performed by using the semantic segmentation of the scene, but this is not implemented as of yet.

\section{Generated Dataset}
\begin{figure}[h]%
    \centering
    \subfloat{{\includegraphics[width=5cm]{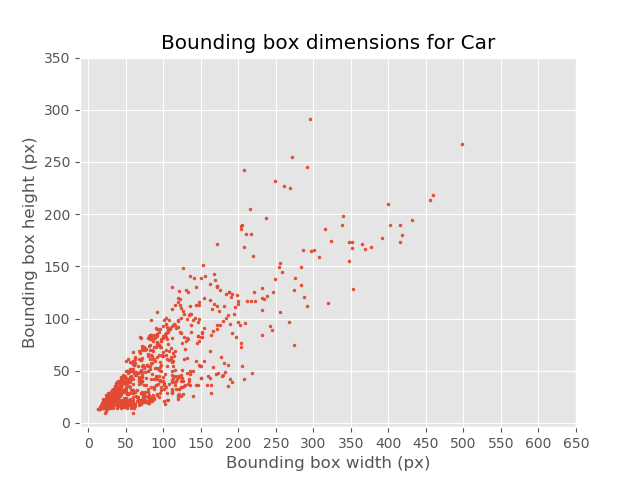} }}%
    \qquad
    \subfloat{{\includegraphics[width=5cm]{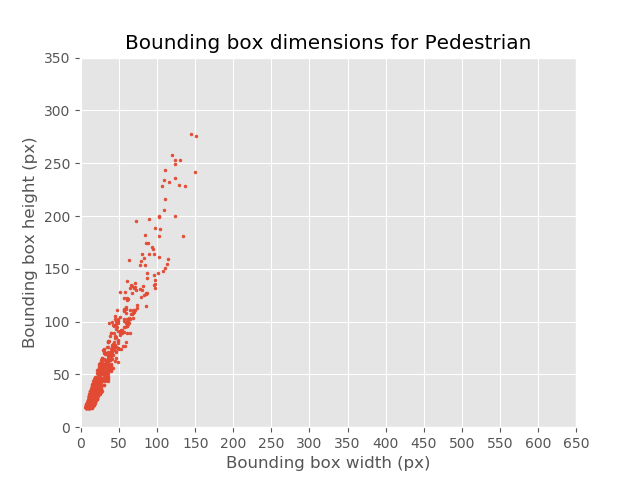} }}%
    \caption{Dimension of 2D bounding boxes for the classes in the CADET dataset.}%
    \label{fig:bbox_dimension_distribution}%
\end{figure}

\begin{figure}[h]%
    \centering
    \subfloat{{\includegraphics[width=5cm]{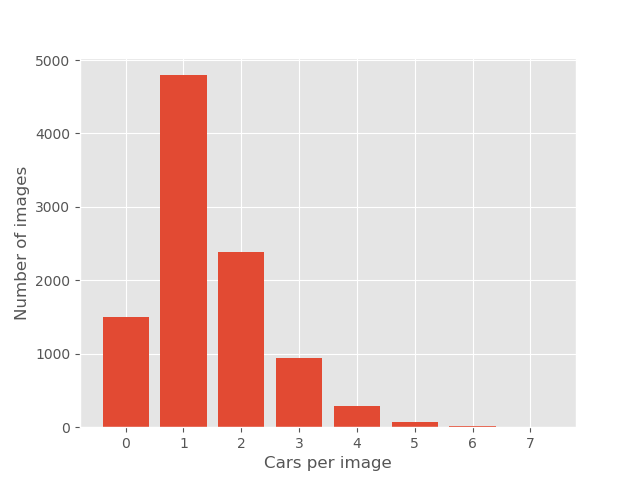} }}%
    \qquad
    \subfloat{{\includegraphics[width=5cm]{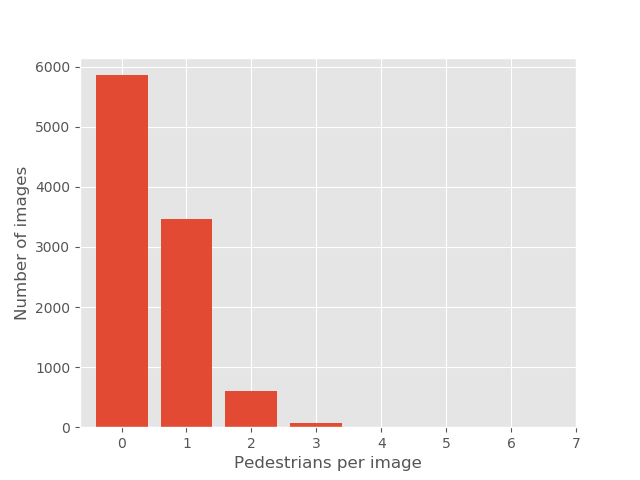} }}%
    \caption{Number of annotations for each class per image in the CADET dataset.}%
    \label{fig:class_distribution}%
\end{figure}

\begin{figure}[h]%
    \centering
    \subfloat{{\includegraphics[width=5cm]{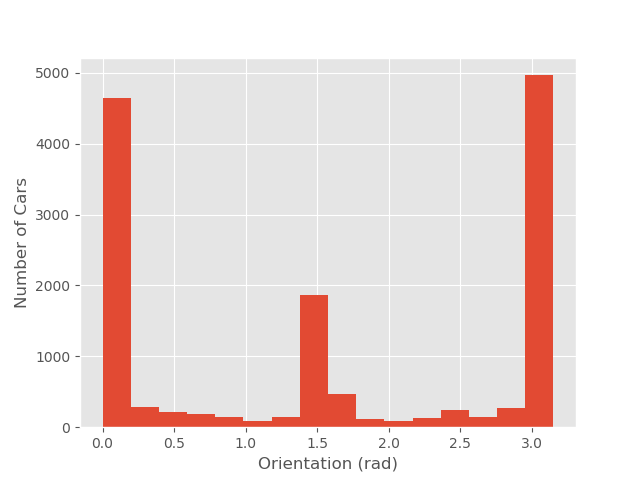}}}%
    \qquad
    \subfloat{{\includegraphics[width=5cm]{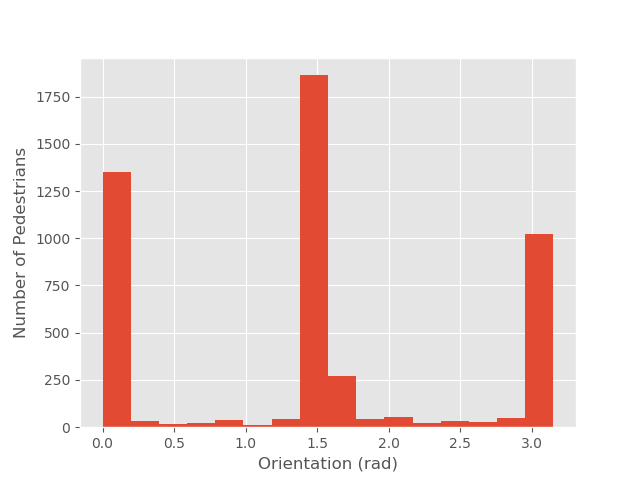} }}%
    \caption{Distribution of orientation per class in the CADET dataset.}%
    \label{fig:orientation_distribution}%
\end{figure}

Using CADET we generate the CADET dataset, consisting of 10 000 samples. In total, there are 13989 cars and 4895 pedestrians in the dataset, averaging about 1.9 labeled objects per image. The dataset contains 2D and 3D bounding box annotations of the classes Car and Pedestrian, and contains both LIDAR and camera sensor data, as well as ground plane estimation and generation of sensor calibration matrices. The environment is generated from two maps, namely \textit{Town01} and \textit{Town02} in the CARLA simulator, which are both suburban environments. The distribution of objects in each image is shown in Figure \ref{fig:class_distribution}. In comparison with the KITTI dataset, the CADET dataset has less cars and pedestrians per image, which is mostly due to city-environment of KITTI, where cars are frequently parked at the side of the road and pedestrians are present in a higher degree. The orientation of each labeled object is shown in Figure \ref{fig:orientation_distribution}. We observe that the distribution of orientation have a sharp multimodal distribution with three peaks, namely for objects seen from the front, behind or sideways. Note that the pedestrians in the dataset generally have a smaller bounding box than cars, shown in Figure \ref{fig:bbox_dimension_distribution}, making them harder to detect.

\section{Training on Simulated Data}
In order to evaluate the use of the simulated CADET dataset, as well experiment with LIDAR feature map representations, several configurations were used of the AVOD-FPN~\cite{avod-fpn} architecture for 3D object detection using camera and LIDAR point cloud. The AVOD-FPN source code has been altered to allow for customized configurations by specifying the features wanted for two groups, slice maps and cloud maps. Slice maps refer to feature maps taken from each vertical slice the point cloud is split into, as specified in the configuration files, while the cloud maps consider the whole point cloud.
Following the approach described in \cite{avod-fpn}, two networks were used for detecting cars and pedestrians separately, repeating the process for each configuration. As multiclass detection might also produce more unstable results when evaluating per class, this was considered the better option. All models used a feature pyramid network to extract features from images and LIDAR, with early fusion of the extracted camera and LIDAR features. Training data is augmented using flipping and jitter, with the only differences between models of the same class being the respective representations of LIDAR feature maps in Bird's Eye View (BEV) as described in Section \ref{model_config}. All configurations used are available in the source code~\cite{avod_fork_github}.

\subsection{Model Configurations} \label{model_config}
\begin{figure}[!h]
	\centering
	\begin{minipage}[t]{0.45\textwidth}
        \centering
        \includegraphics[width=\textwidth]{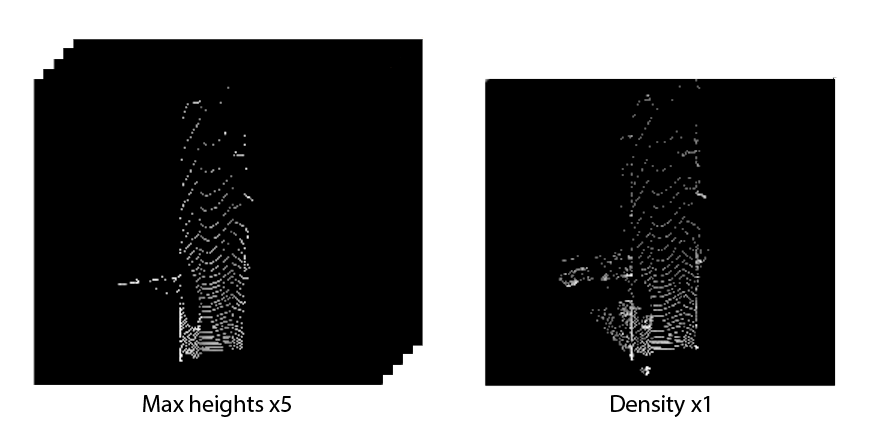}
        \caption{Visualization of default BEV configuration, taking the maximum height within 5 vertical slices as well as the density of the full point cloud.}
        \label{fig:avod-default}
	\end{minipage}
	\hspace{1cm}
	\begin{minipage}[t]{0.45\textwidth}
        \centering
        \includegraphics[width=\textwidth]{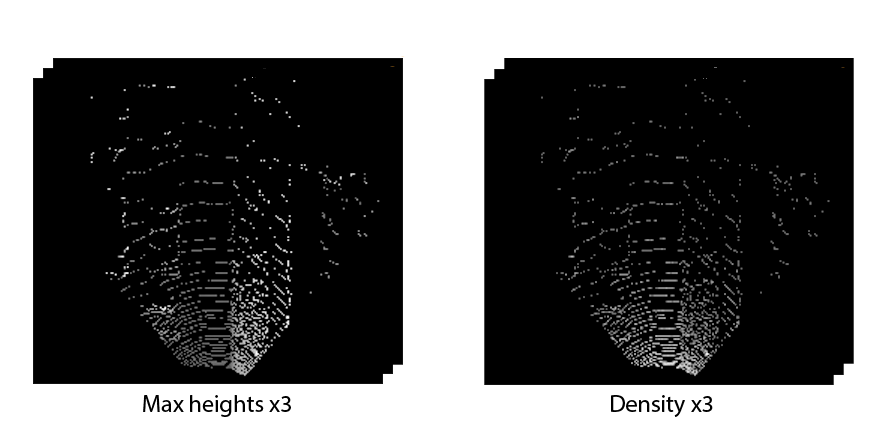}
        \caption{Visualization of first custom BEV configuration, taking the maximum height and density within 3 vertical slices.}
        \label{fig:avod-custom-1}
	\end{minipage}
\end{figure}

\begin{figure}[!h]
    \centering
    \includegraphics[width=0.75\textwidth]{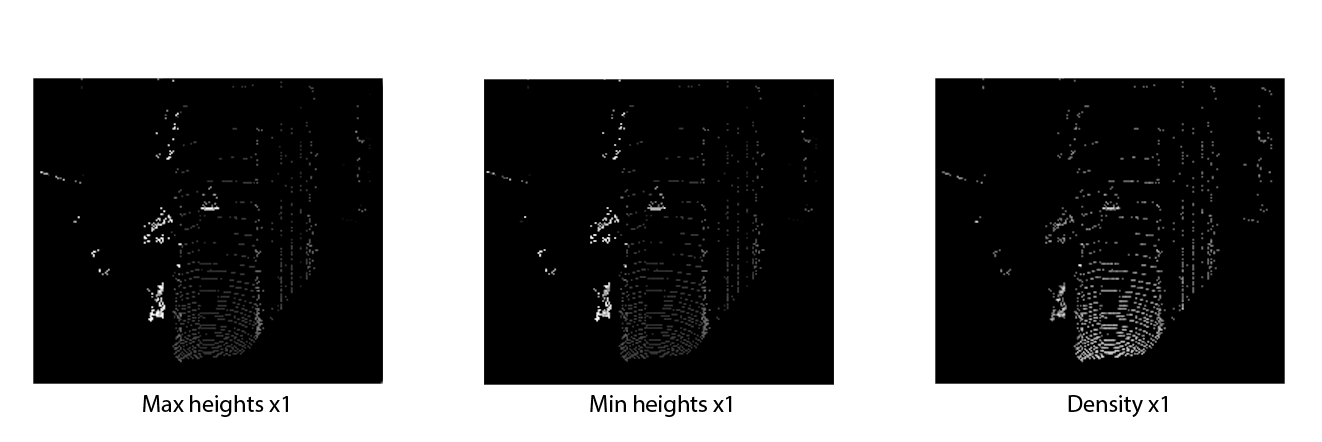}
    \caption{Visualization of second custom BEV configuration, taking the maximum height, minimum height and density of the full point cloud.}
    \label{fig:avod-custom-2}
\end{figure}

AVOD-FPN uses a simplified feature extractor based on the VGG-16 architecture~\cite{vgg} to produce feature maps from camera view as well as LIDAR projected to BEV, allowing the LIDAR to be processed by a Convolutional Neural Network (CNN) designed for 2D images. These separate feature maps are fused together using trainable weights, allowing the model to learn how to best combine multimodal information. In addition to what will be referred to as the default BEV configuration, as proposed in~\cite{avod-fpn}, two additional novel configurations are proposed for which experimental results either show faster inference with similar accuracy, or better accuracy with similar inference speed. In all cases the BEV is discretized horizontally into cells at a resolution of 0.1m. The default configuration creates 5 equally sized vertical slices within a specified height range, taking the highest point in each cell normalized by the slice height. A separate image for the density of the entire point cloud is generated from the number of points $N$ in each cell following Equation \ref{eq:density}, as used in~\cite{avod-fpn} and~\cite{mv3d}, though normalized by $log(64)$ in the latter. We propose a simplified structure, taking the global maximum height, minimum height and density of each cell over the entire point cloud, avoiding the use of slices and halving the amount of BEV maps. We argue that this is sufficient to determine which points belong to large objects and which are outliers, and that it sufficiently defines box dimensions. For classes that occupy less space, we argue that taking three slices vertically using the maximum height and density for each slice can perform better with less susceptibility to noise, as the network could potentially learn to distinguish whether the maximum height value of a slice belongs to the object or not depending on the slice density. All configurations are visualized in Figures \ref{fig:avod-default}-\ref{fig:avod-custom-2}.

\vspace{-2em}

\begin{equation}
    min(1.0, \frac{log(N+1)}{log(16)})
    \label{eq:density}
\end{equation}

\subsection{Results}
In order to gather qualitative results each model trained for a total of 120k steps on the respective datasets, with a batch size of 1, as described in~\cite{avod-fpn}. Checkpoints were stored at every 2k steps, of which the last 20 were selected for evaluation. Tables \ref{table:avod_results_cars} and \ref{table:avod_results_pedestrians} show generated results on the KITTI dataset, for the Car and Pedestrian classes respectively, selecting the best performing checkpoint for each of the 3 BEV configurations. To measure inference speed, each model performs inference on the first 2000 images of the validation set, with learning deactivated, using a NVIDIA GTX 1080 graphics card. The mean inference time is rounded up to the nearest millisecond and presented in the tables.

\begin{table}[h!]
\centering
\caption{KITTI-trained model evaluated on the KITTI dataset for the Car class}
\label{table:avod_results_cars}
\begin{adjustbox}{width=\textwidth}
\def\arraystretch{1.5}%
\setlength{\tabcolsep}{10pt} 
\begin{tabular}{ c c c c c c c c}
& &\multicolumn{3}{c}{\large{$AP_{3D} (\%)$}} & \multicolumn{3}{c}{\large{$AP_{BEV} (\%)$}}  \\ \hline
 \multicolumn{1}{c}{Method} & \multicolumn{1}{c|}{Runtime (ms)} &\multicolumn{1}{c}{\textbf{Easy}} & \multicolumn{1}{c}{\textbf{Moderate}} & \multicolumn{1}{c|}{\textbf{Hard}} & \multicolumn{1}{c}{\textbf{Easy}} & \multicolumn{1}{c}{\textbf{Moderate}} & \multicolumn{1}{c}{\textbf{Hard}} \\ \hline
Default             & 119 & \textbf{83.46} & 73.94 & 67.81 & 89.37 & 86.44 & 78.64 \\
Max*3, Density*3    & 120 & 83.16 & \textbf{73.97} & \textbf{67.98} & \textbf{89.84} & \textbf{86.62} & \textbf{79.85} \\
Max, Min, Density   & \textbf{114} & 82.98 & 73.92 & 67.84 & 89.62 & 86.61 & 79.68 \\
\end{tabular}
\end{adjustbox}
\end{table}

\begin{table}[h!]
\centering
\caption{KITTI-trained model evaluated on the KITTI dataset for the Pedestrian class}
\label{table:avod_results_pedestrians}
\begin{adjustbox}{width=\textwidth}
\def\arraystretch{1.5}%
\setlength{\tabcolsep}{10pt} 
\begin{tabular}{ c c c c c c c c}
& &\multicolumn{3}{c}{\large{$AP_{3D} (\%)$}} & \multicolumn{3}{c}{\large{$AP_{BEV} (\%)$}}  \\ \hline
 \multicolumn{1}{c}{Method} & \multicolumn{1}{c|}{Runtime (ms)} &\multicolumn{1}{c}{\textbf{Easy}} & \multicolumn{1}{c}{\textbf{Moderate}} & \multicolumn{1}{c|}{\textbf{Hard}} & \multicolumn{1}{c}{\textbf{Easy}} & \multicolumn{1}{c}{\textbf{Moderate}} & \multicolumn{1}{c}{\textbf{Hard}} \\ \hline
Default             & 122 & 41.05 & 37.00 & 32.00 & 44.12 & 39.54 & 38.11 \\
Max*3, Density*3    & 122 & \textbf{45.61} & \textbf{42.66} & \textbf{38.06} & \textbf{49.16} & \textbf{45.99} & \textbf{44.53} \\
Max, Min, Density   & \textbf{117} & 27.85 & 27.17 & 24.54 & 33.39 & 33.10 & 29.78 \\
\end{tabular}
\end{adjustbox}
\end{table}

Following evaluation on the KITTI dataset, all configurations were trained from scratch on the generated CADET dataset following the exact same process. Results from evaluation on the validation set of the CADET dataset can be seen in Tables \ref{table:avod_results_cars_carla} and \ref{table:avod_results_pedestrians_carla}. Note that as dynamic occlusion and truncation measurements are not included in the dataset (these are only used for post training evaluation in KITTI), evaluation does not follow the regular easy, moderate, hard categories used in KITTI. Instead objects are categorized as large or small, following the minimum height requirements for the bounding boxes of 40 pixels for easy and 25 pixels for moderate and hard. These models were additionally evaluated directly on the KITTI validation set, with results summarized in Tables \ref{table:avod_results_cars_kitti} and \ref{table:avod_results_pedestrians_kitti}.

\begin{table}[h!]
\centering
\caption{CADET-trained model evaluated on the CADET dataset for the Car class}
\label{table:avod_results_cars_carla}
\begin{adjustbox}{width=0.7\textwidth}
\def\arraystretch{1.5}%
\setlength{\tabcolsep}{10pt} 
\begin{tabular}{ c c c c c }
&\multicolumn{2}{c}{\large{$AP_{3D} (\%)$}} & \multicolumn{2}{c}{\large{$AP_{BEV} (\%)$}}  \\ \hline
 \multicolumn{1}{c|}{Method} &\multicolumn{1}{c}{\textbf{Large}} & \multicolumn{1}{c|}{\textbf{Small}} & \multicolumn{1}{c}{\textbf{Large}} & \multicolumn{1}{c}{\textbf{Small}} \\ \hline
Default             & 70.86 & 69.37 & \textbf{80.13} & \textbf{71.32} \\
Max*3, Density*3    & \textbf{70.96} & \textbf{69.59} & 79.81 & 71.28 \\
Max, Min, Density   & 68.79 & 60.87 & 78.75 & 70.72 \\
\end{tabular}
\end{adjustbox}
\end{table}

\begin{table}[h!]
\centering
\caption{CADET-trained model evaluated on the CADET dataset for the Pedestrian class}
\label{table:avod_results_pedestrians_carla}
\begin{adjustbox}{width=0.7\textwidth}
\def\arraystretch{1.5}%
\setlength{\tabcolsep}{10pt} 
\begin{tabular}{ c c c c c }
&\multicolumn{2}{c}{\large{$AP_{3D} (\%)$}} & \multicolumn{2}{c}{\large{$AP_{BEV} (\%)$}}  \\ \hline
 \multicolumn{1}{c|}{Method} &\multicolumn{1}{c}{\textbf{Large}} & \multicolumn{1}{c|}{\textbf{Small}} & \multicolumn{1}{c}{\textbf{Large}} & \multicolumn{1}{c}{\textbf{Small}} \\ \hline
Default             & 75.43 & \textbf{73.89} & 75.43 & \textbf{73.91} \\
Max*3, Density*3    & \textbf{76.13} & 72.99 & \textbf{80.24} & 73.41 \\
Max, Min, Density   & 75.49 & 71.82 & 79.73 & 72.37 \\
\end{tabular}
\end{adjustbox}
\end{table}

\begin{table}[h!]
\centering
\caption{CADET-trained model evaluated on the KITTI dataset for the Car class}
\label{table:avod_results_cars_kitti}
\begin{adjustbox}{width=0.9\textwidth}
\def\arraystretch{1.5}%
\setlength{\tabcolsep}{10pt} 
\begin{tabular}{ c c c c c c c }
&\multicolumn{3}{c}{\large{$AP_{3D} (\%)$}} & \multicolumn{3}{c}{\large{$AP_{BEV} (\%)$}}  \\ \hline
 \multicolumn{1}{c|}{Method} &\multicolumn{1}{c}{\textbf{Easy}} & \multicolumn{1}{c}{\textbf{Moderate}} & \multicolumn{1}{c|}{\textbf{Hard}} & \multicolumn{1}{c}{\textbf{Easy}} & \multicolumn{1}{c}{\textbf{Moderate}} & \multicolumn{1}{c}{\textbf{Hard}} \\ \hline
Default             & 29.85 & 20.29 & 18.40 & 50.58 & 37.81 & 30.77 \\
Max*3, Density*3    & \textbf{35.85} & \textbf{29.40} & \textbf{24.99} & \textbf{57.32} & \textbf{49.63} & \textbf{43.25} \\
Max, Min, Density   & 30.34 & 24.28 & 20.25 & 45.22 & 37.56 & 31.17 \\
\end{tabular}
\end{adjustbox}
\end{table}

\begin{table}[h!]
\centering
\caption{CADET-trained model evaluated on the KITTI dataset for the Pedestrian class}
\label{table:avod_results_pedestrians_kitti}
\begin{adjustbox}{width=0.9\textwidth}
\def\arraystretch{1.5}%
\setlength{\tabcolsep}{10pt} 
\begin{tabular}{ c c c c c c c }
&\multicolumn{3}{c}{\large{$AP_{3D} (\%)$}} & \multicolumn{3}{c}{\large{$AP_{BEV} (\%)$}}  \\ \hline
 \multicolumn{1}{c|}{Method} &\multicolumn{1}{c}{\textbf{Easy}} & \multicolumn{1}{c}{\textbf{Moderate}} & \multicolumn{1}{c|}{\textbf{Hard}} & \multicolumn{1}{c}{\textbf{Easy}} & \multicolumn{1}{c}{\textbf{Moderate}} & \multicolumn{1}{c}{\textbf{Hard}} \\ \hline
Default             & \textbf{9.09} & \textbf{9.09} & \textbf{9.09} & 9.38 & \textbf{9.33} & \textbf{9.38} \\
Max*3, Density*3    & 2.27 & 2.27 & 2.27 & 2.27 & 2.27 & 2.27 \\
Max, Min, Density   & \textbf{9.09} & \textbf{9.09} & \textbf{9.09} & \textbf{9.78} & 9.09 & 9.09 \\
\end{tabular}
\end{adjustbox}
\end{table}

The CADET-trained models were subsequently restored from their checkpoints at step 90k and modified for further training on the KITTI dataset. Training was resumed until step 150k, meaning the models received 60k steps of training on the KITTI training set as opposed to 120k originally. Other than increasing the amount of steps and switching the target datasets, the configuration files were not altered from when training on the CADET dataset. Tables \ref{table:avod_results_cars_finetune} and \ref{table:avod_results_pedestrians_finetune} show results from the top performing checkpoint of each model.

\begin{table}[h!]
\centering
\caption{CADET-trained model, fine-tuned and evaluated on the KITTI dataset for the Car class}
\label{table:avod_results_cars_finetune}
\begin{adjustbox}{width=0.9\textwidth}
\def\arraystretch{1.5}%
\setlength{\tabcolsep}{10pt} 
\begin{tabular}{ c c c c c c c }
&\multicolumn{3}{c}{\large{$AP_{3D} (\%)$}} & \multicolumn{3}{c}{\large{$AP_{BEV} (\%)$}}  \\ \hline
 \multicolumn{1}{c|}{Method} &\multicolumn{1}{c}{\textbf{Easy}} & \multicolumn{1}{c}{\textbf{Moderate}} & \multicolumn{1}{c|}{\textbf{Hard}} & \multicolumn{1}{c}{\textbf{Easy}} & \multicolumn{1}{c}{\textbf{Moderate}} & \multicolumn{1}{c}{\textbf{Hard}} \\ \hline
Default             & \textbf{83.84} & 68.67 & \textbf{67.40} & \textbf{89.41} & 79.77 & \textbf{78.86} \\
Max*3, Density*3    & 76.85 & \textbf{72.44} & 66.55 & 88.20 & \textbf{85.18} & 78.71 \\
Max, Min, Density   & 81.00 & 66.95 & 65.88 & 88.76 & 79.36 & 78.41 \\
\end{tabular}
\end{adjustbox}
\end{table}

\vspace{-2em}
\begin{table}[h!]
\centering
\caption{CADET-trained model, fine-tuned and evaluated on the KITTI dataset for the Pedestrian class}
\label{table:avod_results_pedestrians_finetune}
\begin{adjustbox}{width=0.9\textwidth}
\def\arraystretch{1.5}%
\setlength{\tabcolsep}{10pt} 
\begin{tabular}{ c c c c c c c }
&\multicolumn{3}{c}{\large{$AP_{3D} (\%)$}} & \multicolumn{3}{c}{\large{$AP_{BEV} (\%)$}}  \\ \hline
 \multicolumn{1}{c|}{Method} &\multicolumn{1}{c}{\textbf{Easy}} & \multicolumn{1}{c}{\textbf{Moderate}} & \multicolumn{1}{c|}{\textbf{Hard}} & \multicolumn{1}{c}{\textbf{Easy}} & \multicolumn{1}{c}{\textbf{Moderate}} & \multicolumn{1}{c}{\textbf{Hard}} \\ \hline
Default             & \textbf{40.26} & \textbf{38.55} & 33.93 & \textbf{46.96} & \textbf{44.80} & \textbf{40.73} \\
Max*3, Density*3    & 39.19 & 38.02 & \textbf{34.15} & 46.14 & 43.54 & 40.44 \\
Max, Min, Density   & 37.32 & 34.34 & 32.60 & 45.71 & 42.43 & 37.62 \\
\end{tabular}
\end{adjustbox}
\end{table}

\section{Discussion}
For the fully KITTI-trained models, results on the Car class are very similar for all configurations, where the largest loss in amounts to only 0.5\% 3D AP on the easy category from the default configuration to our configuration using half as many layers in the BEV map. Larger differences are apparent for the Pedestrian class, where 3 layers is not sufficient to compete with the default configuration. However, the use of 3 slices of maximum heights and density, totalling 6 layers as with the default configuration, shows noticeably better results across the board suggesting a more robust behaviour.

Evaluation of the CADET-trained models on the CADET validation set shows similar relative performance between the models, however with the simpler custom configuration showing a dip in accuracy for the moderate category on the car class. With regards to pedestrians, differences are much smaller than what could have been expected. Also considering the unremarkable and rather inconsistent performance on the Pedestrian class of the KITTI dataset, we can likely accredit this to overly simplified representation of the physical collision of pedestrians visible in the simulated LIDAR point cloud. The CADET-trained models does perform better on the car class however, suggesting better and more consistent generalization this task.

The fine-tuned models show performance on the Car class mostly similar to the fully KITTI-trained models, however with each model showing a noticeable drop in performance on either the easy or moderate category. Results on the Pedestrian class are a bit more interesting. The default configuration sees a slight increase in accuracy on the moderate and hard category, with a slight decrease for easy. The max/density configuration sees a significant decrease in performance on all categories, where as the less complex max/min/density configuration, although still being the weakest performer, sees a significant increase in performance compared to when only trained on the KITTI dataset. The reason for the rather inconsistent results when compared to the KITTI-trained models are not thoroughly investigated, but can in part be due to somewhat unstable gradients not producing fully reliable results. The CARLA generated LIDAR point cloud does not feature accurate geometry due to simplified collision of all dynamic objects. As such the different capabilities of the configurations may not be exploited during the pretraining on the CADET dataset, impacting overall results. The simplest configuration significantly closing the gap on the Pedestrian class may be a testament to this, as the simplified pedestrian representation is more easily recognizable.

While results from simulated and partly simulated training do not generally exceed the performance of direct training on the dataset, there is a clear indication that the use of simulated data can achieve closely matched performance with less training on actual data. The ease of generation and expandabililty in terms of sensors, scenarios, environments and conditions makes tools such as CADET very useful for training and evaluating models for autonomous driving, although improvements are needed before they are sufficient for training real world solutions. 

\section{Conclusion}
In recent years, the use of synthetic data for training machine learning models has gained in popularity due to the costs associated with gathering real-life data. This is especially true with regards to autonomous driving because of the strict demand of generalizability to a diverse number of driving scenarios. In this study, we have described CADET - a tool for generating large amounts of training data for perception in autonomous driving, and the resulting dataset. We have demonstrated that this dataset, while not sufficient to directly train systems for use in the real world, is useful in lowering the amount of real-life data required to train machine learning models to reasonably high levels of accuracy. We have also suggested and evaluated two novel BEV representations, easily configurable before training, with potential for better detection of smaller objects and reduced complexity for detection of larger objects respectively. The CADET toolkit, while still requiring improved physical models in LIDAR modelling, is currently able to generate datasets for training and validation of virtually any model designed for the KITTI object detection task.

\renewcommand\floatpagefraction{.9}
\renewcommand\topfraction{.9}
\renewcommand\bottomfraction{.9}
\renewcommand\textfraction{.1} 

%
%
\clearpage
\bibliographystyle{splncs04}

\end{document}